\documentclass[10pt]{article}

\usepackage[letterpaper,left=0.984in,right=0.984in,top=0.79in,bottom=0.72in]{geometry}
\usepackage{newpxtext,newpxmath}
\usepackage{microtype}
\usepackage{amsmath}
\usepackage{graphicx}
\usepackage{booktabs}
\usepackage{multirow}
\usepackage{array}
\usepackage{pifont}
\usepackage[table,dvipsnames]{xcolor}
\usepackage{hyperref}
\usepackage{url}
\usepackage[numbers,sort&compress]{natbib}
\usepackage{enumitem}
\usepackage{xspace}
\usepackage{titlesec}
\usepackage[most]{tcolorbox}

\definecolor{releaseblue}{HTML}{175CCB}
\definecolor{abstractbg}{HTML}{EEF4FF}
\definecolor{oursbg}{HTML}{EAF1FF}
\definecolor{markfull}{HTML}{15803D}
\definecolor{markpart}{HTML}{D97706}
\definecolor{marknone}{HTML}{B8BDC7}
\definecolor{logoblue}{HTML}{3157C8}
\definecolor{logopurple}{HTML}{7A3FA0}
\definecolor{logored}{HTML}{C73755}

\hypersetup{
  colorlinks=true,
  citecolor=releaseblue,
  linkcolor=releaseblue,
  urlcolor=releaseblue
}

\titleformat{\section}
  {\fontsize{13}{15.6}\selectfont\bfseries\color{releaseblue}}
  {\thesection}{0.72em}{}
\titleformat{\subsection}
  {\fontsize{11.2}{13.4}\selectfont\bfseries\color{releaseblue}}
  {\thesubsection}{0.62em}{}
\titleformat{\subsubsection}
  {\normalsize\bfseries\color{releaseblue}}
  {\thesubsubsection}{0.55em}{}
\titlespacing*{\section}{0pt}{1.15em}{0.55em}
\titlespacing*{\subsection}{0pt}{0.85em}{0.35em}
\titlespacing*{\subsubsection}{0pt}{0.72em}{0.28em}

\setlist[itemize]{leftmargin=1.55em,itemsep=2.4pt,topsep=3.5pt}

\DeclareRobustCommand{\liveassistant}{LiveAssistant\xspace}

\newcommand{\cfull}{\textcolor{markfull}{\ensuremath{\bullet}}}
\newcommand{\cpart}{\textcolor{markpart}{\ensuremath{\odot}}}
\newcommand{\cnone}{\textcolor{marknone}{\ensuremath{\circ}}}
\newcommand{\ourstar}{\textcolor{Purple!70!black}{\ding{72}}}

\newtcolorbox{releaseabstract}{
  enhanced,
  colback=abstractbg,
  colframe=abstractbg,
  boxrule=0pt,
  arc=3.2mm,
  outer arc=3.2mm,
  width=\dimexpr\textwidth+24pt\relax,
  enlarge left by=-12pt,
  left=28pt,
  right=28pt,
  top=5.5mm,
  bottom=5.4mm,
  before skip=0pt,
  after skip=0pt
}

\begin{document}

\noindent
\begin{minipage}[c]{0.57\textwidth}
  \includegraphics[width=0.76\linewidth]{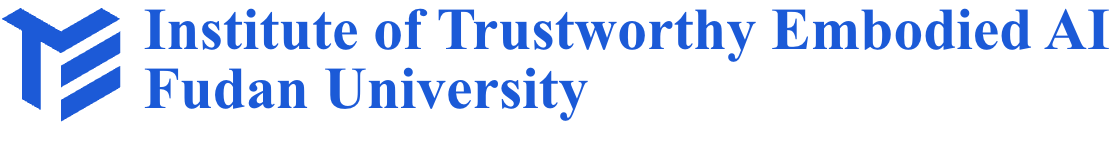}
\end{minipage}%
\begin{minipage}[c]{0.43\textwidth}
  \raggedleft
  \includegraphics[width=0.77\linewidth]{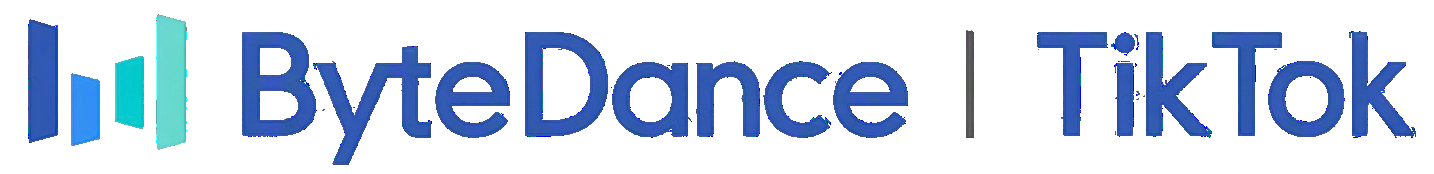}
\end{minipage}

\vspace{0.15em}
{\color{releaseblue}\hrule height 1.15pt}
\vspace{0.62em}

\begin{center}
{\fontsize{18.2}{21.5}\selectfont\bfseries
\makebox[\textwidth][c]{\textsc{Live Assistant}: Learning Whether, When, and Whom to}\\[-0.05em]
\makebox[\textwidth][c]{Assist in Real-World Live Social Streams}\par}
\end{center}

\vspace{0.52em}
{\color{releaseblue}\hrule height 0.85pt}
\vspace{0.12em}

\begin{center}
{\fontsize{12.5}{15.0}\selectfont\bfseries
Shujian Gao\textsuperscript{1,2,3,\#},
Jiamei Yan\textsuperscript{2,\#},
Yuchen Yang\textsuperscript{2,\#},
Penghao Zhou\textsuperscript{2,$\dagger$},
Qinglei Wang\textsuperscript{2,$\dagger$},\\[-0.08em]
Tiehan Fan\textsuperscript{2},
Yuan Wang\textsuperscript{4},
Zuxuan Wu\textsuperscript{1,3,*},
Yu-gang Jiang\textsuperscript{1,*}\par}

\vspace{0.62em}
{\fontsize{10}{12.5}\selectfont
\textsuperscript{1}Institute of Trustworthy Embodied AI, Fudan University,
\textsuperscript{2}ByteDance TikTok,\\[-0.10em]
\textsuperscript{3}Shanghai Innovation Institution\\[-0.10em]
\textsuperscript{4}Zhejiang University\par}
\end{center}

\vspace{1.95em}

\begin{releaseabstract}
\begin{center}
{\fontsize{12}{14.5}\selectfont\bfseries\color{releaseblue}Abstract}
\end{center}
\vspace{-0.28em}

Livestreams are long lasting interactive environments in which audio visual content, viewer activity, host behavior, and platform signals evolve together. This setting creates a streaming assistance problem whose requests emerge from the environment itself. We introduce \liveassistant, a framework for mixed initiative and role conditioned assistance that formulates livestream interaction as four coupled decisions: \textbf{whether to act, when to act, whom to address, and what to communicate}. The model continuously consumes native audio and video together with synchronized comments, gifts, viewer dynamics, and room metadata. At every ten second interval, one autoregressive policy selects \textsc{OBS}, \textsc{MEM}, or \textsc{ANS}. \textsc{OBS} preserves deliberate silence, \textsc{MEM} records a private semantic update, and \textsc{ANS} specifies a recipient, a task, and a grounded message for viewers or the host. This unified protocol makes activation, routing, and content generation depend on the same evolving stream representation. To support the task, we construct a trajectory data engine that reconstructs real livestream sessions, aligns heterogeneous platform events with native media, and produces structured causal supervision. The resulting optimization corpus contains more than 320 hours of trajectories, while the held out benchmark contains 275 complete clips and 13,812 unique decision intervals with human review. We train the policy in two stages. Marker Aware Multiturn Supervised Fine Tuning, MA-MSFT, strengthens sparse state, recipient, and task decisions inside open vocabulary generation. Streaming Multiturn GSPO, SM-GSPO, then optimizes self generated trajectories with scoped turn level and trajectory level credit. On the held out benchmark, \liveassistant reaches 71.14 state accuracy, 72.67 recipient accuracy, and 58.41 task accuracy, with consistent gains over representative streaming video and general multimodal baselines. The benchmark also exposes class specific behavior for silence, memory, and recipient routing. Together, the formulation, benchmark, and training framework establish livestream assistance as selective participation in a shared social stream.

\vspace{2.12em}
{\small\bfseries Correspondence:} {\small\texttt{zxwu@fudan.edu.cn, ygj@fudan.edu.cn}}\\[-0.08em]
{\small\bfseries Website:} {\small\url{https://daryl-gsj.github.io/LiveAssistant/}}
\end{releaseabstract}

\footnotetext{\textsuperscript{*}Corresponding. \textsuperscript{$\dagger$}Project Lead. \textsuperscript{\#}Equal Contribution.}

\vspace{0.83em}
\section{Introduction}
\label{sec:introduction}

Streaming video understanding requires a model to perceive and reason over content as it unfolds. Recent systems have advanced causal perception, online question answering, response timing, native audio visual modeling, and long context reasoning~\citep{chen2024videollmonline,huang2025ovbench,li2025ovobench,qian2025dispider,wang2025streambridge,yang2025livestar}. These capabilities provide the foundation for continuous interaction with video.

Livestreaming extends this setting into a shared social environment. Audio visual content evolves together with comments, gifts, audience dynamics, and room state~\citep{qu2026kuailive,guo2026kuailivem3,wang2026livibench}. The same stream also serves participants with different information needs. Viewers may need an explanation or an answer to a repeated question. Hosts may need cues about audience confusion, pacing, or emerging demand. Platform activity can indicate moments that deserve attention. An assistant therefore has to infer an actionable need from the evolving stream and resolve four decisions together: \textbf{whether} the context calls for action, \textbf{when} the action is useful, \textbf{whom} the action should serve, and \textbf{what} information should be communicated.

Existing research provides complementary parts of this capability. Streaming video models learn causal perception and response timing from an external query, instruction, or trigger~\citep{wang2025streambridge,yang2025livestar,wang2025mmduet2,tian2026roma,xie2026streamov,xu2026streamingvlm}. Livestream systems use comments, gifts, and room dynamics for generation, retrieval, recommendation, and moderation~\citep{ma2019livebot,lalanne2023livechat,wang2026livibench,qu2026kuailive,tang2021videomoderator,yew2026dynamicmoderation}. Multi party interaction research studies initiative and addressee selection in conversational settings~\citep{horvitz1999principles,allen1999mixedinitiative,houde2025groupagent,inoue2025addressee}. Livestream assistance connects these directions through one sequential decision: discover a need from synchronized media and social evidence, place an intervention at a useful moment, route it to an appropriate participant, and produce grounded content.

\begin{figure}[t]
    \centering
    \includegraphics[width=\linewidth]{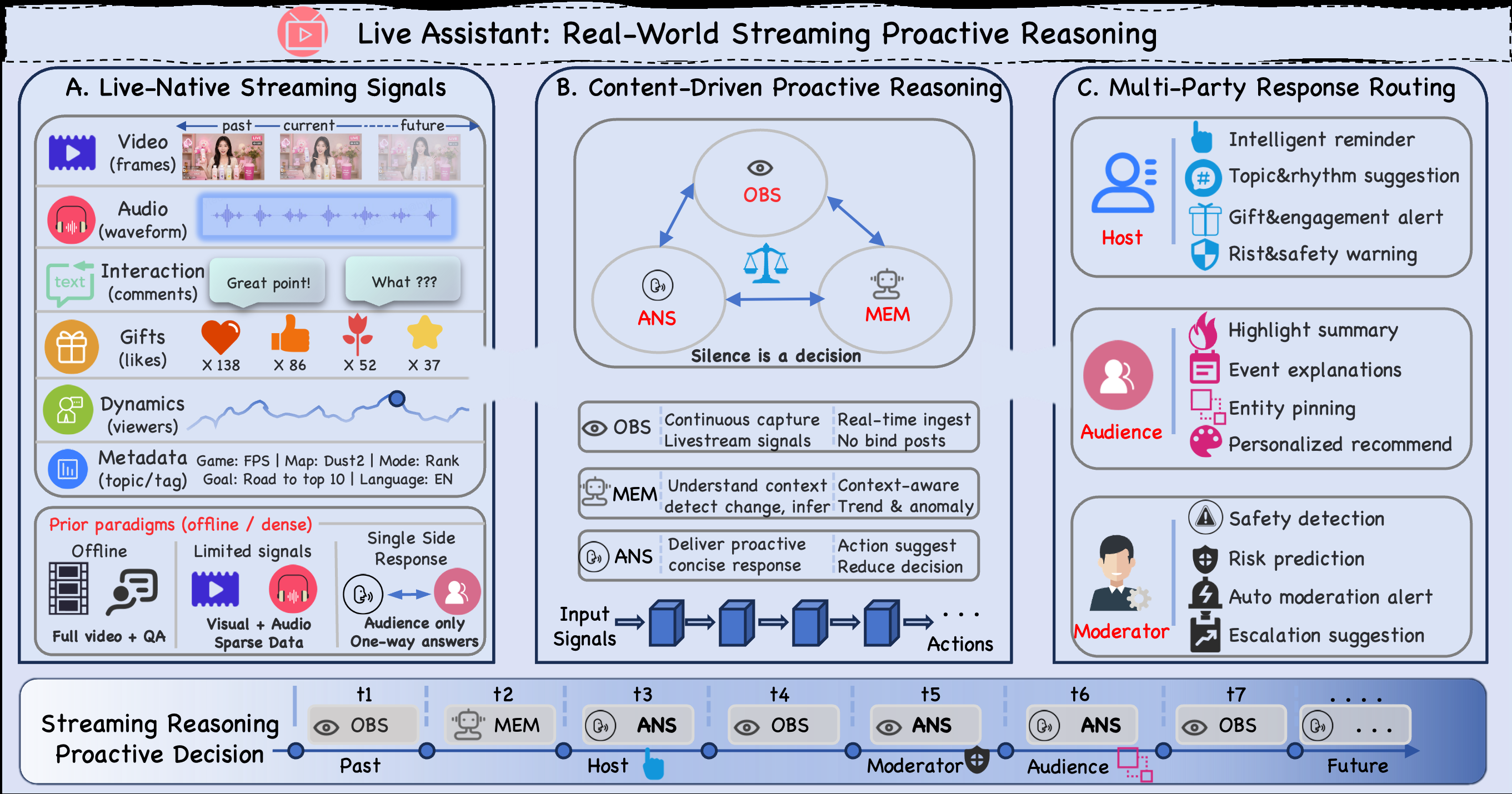}
    \caption{\liveassistant treats a livestream as a heterogeneous social multimodal stream. It learns whether to act, which party should receive the action, and what information should be communicated. The current benchmark evaluates viewer and host assistance.}
    \label{fig:overview}
\end{figure}

We introduce \liveassistant, a framework for mixed initiative, role conditioned assistance in livestreams. At each step, the model observes native audio and video together with synchronized comments, gifts, viewer dynamics, and room metadata. A single autoregressive policy chooses among three actions: continue observing, update private semantic memory, or produce an external response. External responses specify a recipient, a task, and grounded message content. The current benchmark evaluates viewer and host assistance. Figure~\ref{fig:overview} illustrates the broader live room setting and the three signal families used by the model.

Learning this policy introduces three challenges. First, suitable supervision requires time aligned media, platform events, and per interval assistant decisions. We build a trajectory engine that reconstructs real livestreams, aligns synchronized signals, and converts continuous sessions into structured decision trajectories. Second, the decisive state, recipient, and task markers occupy only a small fraction of each generated target. We introduce \textbf{Marker Aware Multiturn Supervised Fine Tuning}, \textbf{MA-MSFT}, which increases supervision on these structural decisions while preserving open vocabulary generation. Third, every model action becomes part of the context for later decisions. We introduce \textbf{Streaming Multiturn GSPO}, \textbf{SM-GSPO}, which trains on self generated trajectories and combines turn level credit with trajectory level credit~\citep{shao2025gspo}.

Our contributions are threefold.
\begin{itemize}
    \item We formulate livestream assistance as a mixed initiative decision problem in which one policy jointly resolves whether to act, when to act, whom to address, and what to communicate from a continuously evolving social stream.
    \item We construct structured livestream trajectories from native audio, video, comments, and platform signals, together with a human reviewed benchmark for action timing, recipient routing, task selection, and grounded response generation from causal stream prefixes.
    \item We develop MA-MSFT and SM-GSPO for structured streaming trajectories. MA-MSFT strengthens sparse decision supervision, and SM-GSPO learns from self generated trajectories with scoped turn and trajectory credit.
\end{itemize}

\section{Problem Formulation}
\label{sec:formulation}

\begin{figure}[t]
    \centering
    \includegraphics[width=\linewidth]{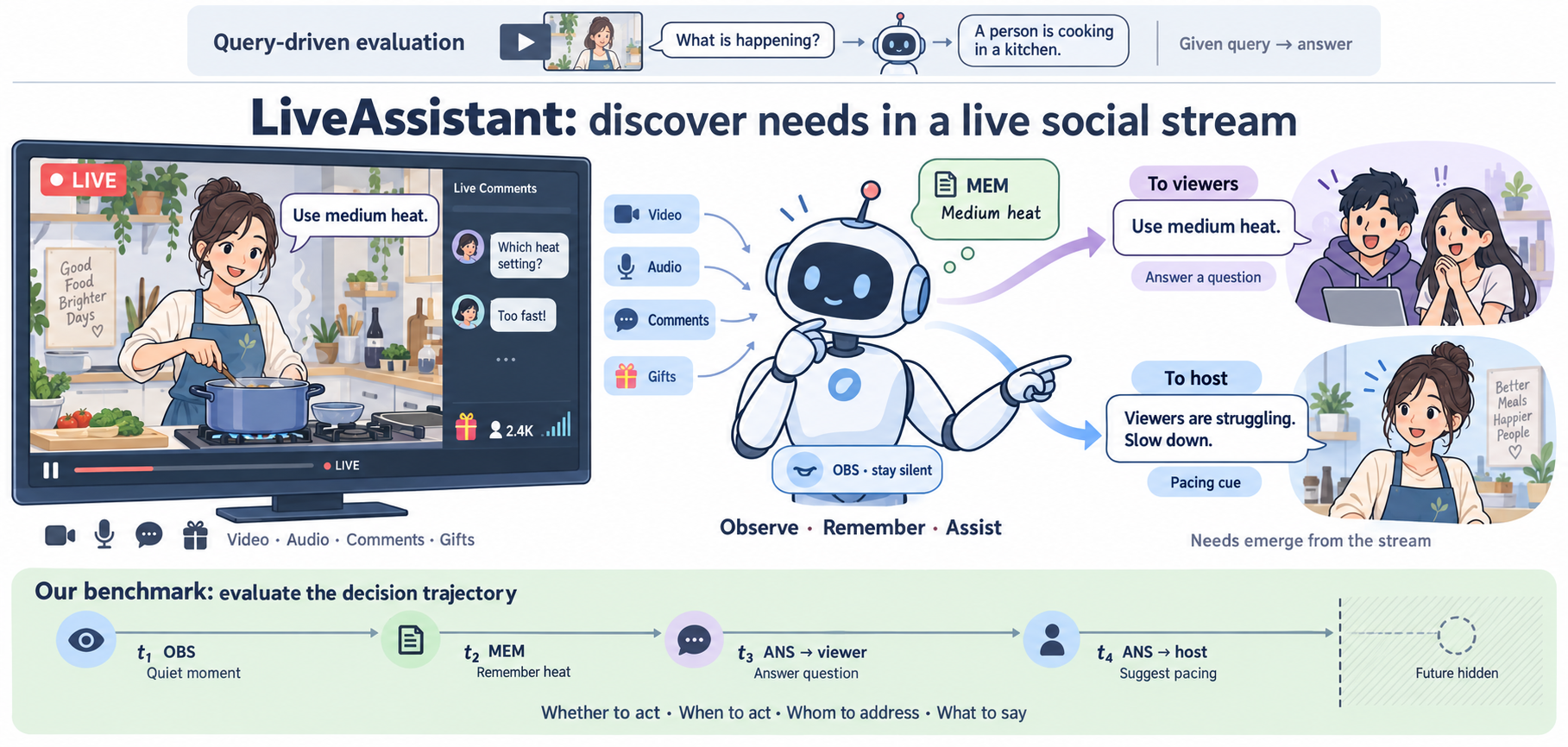}
    \caption{\liveassistant discovers assistance needs from an unfolding stream of audio, video, comments, and platform signals. The policy may remain silent with \textsc{OBS}, update private memory with \textsc{MEM}, or emit an \textsc{ANS} action to viewers or the host. Evaluation follows the complete decision trajectory under causal stream prefixes.}
    \label{fig:problem}
\end{figure}

\subsection{Livestream Assistance as Mixed Initiative Decision Making}

A livestream unfolds as one causal process in which content, audience reactions, and platform state evolve together. Participants share this context while requiring different information at different moments. The assistant receives a persistent role instruction and the observed stream prefix. Assistance needs can arise from media events, repeated viewer requests, changes in audience behavior, or host facing operational cues.

Each decision interval poses four coupled questions. \textbf{Whether to act} identifies a useful opportunity for intervention. \textbf{When to act} places the intervention inside its useful temporal window. \textbf{Whom to address} routes the action to the participant who can use it. \textbf{What to communicate} produces concise grounded content for that participant. A single policy resolves the four decisions from a shared causal representation.

\begin{figure*}[t]
    \centering
    \includegraphics[width=\textwidth]{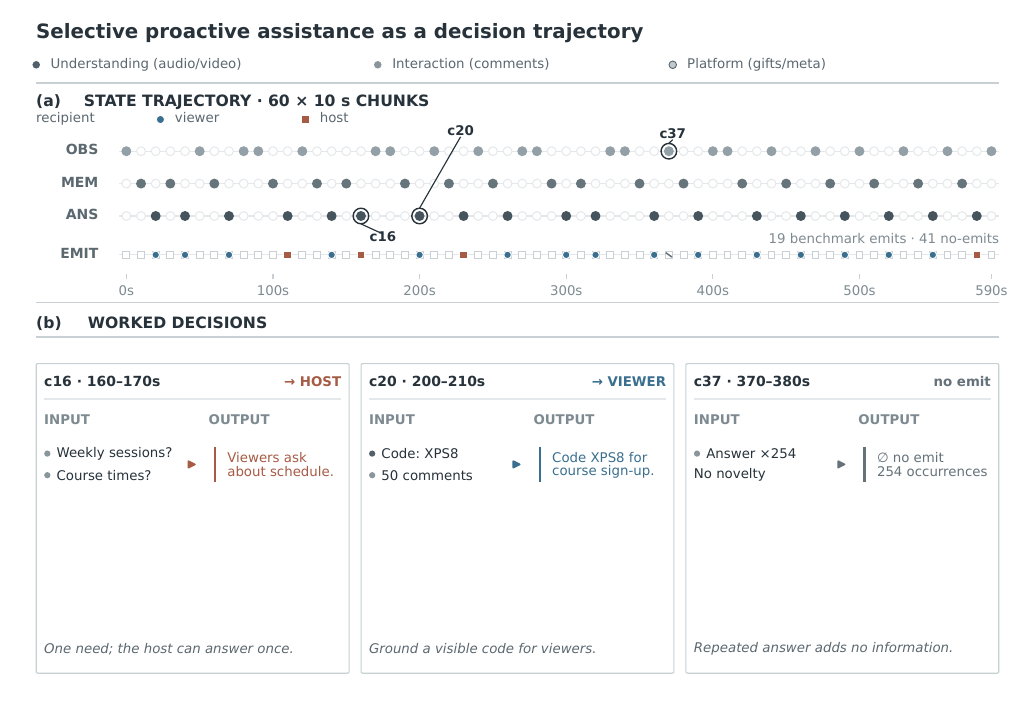}
    \caption{\textbf{Selective assistance over a continuous livestream.}
    \liveassistant processes synchronized media, comments, and platform signals across 60 consecutive intervals. It accumulates relevant context through \textsc{MEM}, routes useful information to viewers or hosts through \textsc{ANS}, and remains in \textsc{OBS} when an intervention provides no additional value.}
    \label{fig:selective_decision_trajectory}
    \vspace{-0.6em}
\end{figure*}

\subsection{Formalization}

A livestream is represented as an ordered sequence
\begin{equation}
    \mathcal{X}=\{x_t\}_{t=1}^{T},
    \qquad
    x_t=(v_t,a_t,e_t),
\end{equation}
where $v_t$ and $a_t$ denote native video and audio, and $e_t$ contains synchronized social signals, including comments, gifts, viewer dynamics, and room metadata. At step $t$, the model receives the available causal history
\begin{equation}
    h_t=(p,x_{\leq t},y_{<t}),
\end{equation}
where $p$ is the persistent instruction and $y_i$ is the assistant action at step $i$.

Given $h_t$, the policy predicts a state
\begin{equation}
    z_t\in\mathcal{S}=\{\mathrm{OBS},\mathrm{MEM},\mathrm{ANS}\}.
\end{equation}
\textsc{OBS} continues observing. \textsc{MEM} writes a private semantic update $u_t$. \textsc{ANS} emits a set of external messages $\mathcal{M}_t$. Each message is represented as
\begin{equation}
    m_{t,k}=(r_{t,k},q_{t,k},w_{t,k}),
\end{equation}
where $r_{t,k}\in\{\mathrm{viewer},\mathrm{host}\}$ is the recipient, $q_{t,k}$ is the task, and $w_{t,k}$ is concise grounded content. The serialized action is
\begin{equation}
 y_t=
 \begin{cases}
 \langle\mathrm{obs}\rangle,
 & z_t=\mathrm{OBS},\\
 \langle\mathrm{mem}\rangle u_t\langle/\mathrm{mem}\rangle,
 & z_t=\mathrm{MEM},\\
 \langle\mathrm{ans}\rangle
 \{\langle r_{t,k}\rangle[q_{t,k}]\,w_{t,k}\}_{k=1}^{|\mathcal{M}_t|}
 \langle/\mathrm{ans}\rangle,
 & z_t=\mathrm{ANS}.
 \end{cases}
 \label{eq:action_protocol}
\end{equation}
The state sequence determines activation and timing. Each answer further determines recipient, task, and content. As illustrated in \autoref{fig:selective_decision_trajectory}, assistance emerges as a selective decision trajectory in which the policy accumulates context, chooses an appropriate recipient, and emits only when the evolving stream supports a useful action.

\subsection{Challenges from the Formulation}

The formulation introduces three learning requirements. \textbf{Trajectory supervision} must pair native livestream inputs with state, recipient, task, and grounded content at each decision interval. \textbf{Sparse structural decisions} require sufficient optimization signal for short markers inside longer generated sequences. \textbf{Compounding actions} require training under the policy history because an earlier state, memory, or response changes subsequent context. These requirements motivate trajectory construction, MA-MSFT, and SM-GSPO.

\begin{table}[t]
\centering
\caption{Comparison of representative work along the two axes of livestream assistance. Social stream input covers native audio visual media, endogenous comment requests, and platform signals. Mixed initiative output covers whether, when, whom, and what. Full, partial, and absent support are shown by \cfull, \cpart, and \cnone.}
\label{tab:comparison}
\scriptsize
\setlength{\tabcolsep}{3.1pt}
\renewcommand{\arraystretch}{1.08}
\resizebox{\linewidth}{!}{%
\begin{tabular}{@{}lcc ccc cccc@{}}
\toprule
\textbf{Work} & \textbf{Type} & \textbf{Dom.} & \multicolumn{3}{c}{\textbf{Social stream input}} & \multicolumn{4}{c}{\textbf{Mixed initiative output}}\\
\cmidrule(lr){4-6}\cmidrule(lr){7-10}
& & & \textbf{A/V} & \textbf{Req.} & \textbf{Plat.} & \textbf{Whether} & \textbf{When} & \textbf{Whom} & \textbf{What}\\
\midrule
StreamingBench~\citep{lin2024streamingbench} & B & V & \cfull & \cnone & \cnone & \cpart & \cpart & 1 & \cfull\\
OVO-Bench~\citep{li2025ovobench} & B & V & \cpart & \cnone & \cnone & \cnone & \cpart & 1 & \cfull\\
OmniMMI~\citep{wang2025omnimmi} & B+M & V & \cfull & \cnone & \cnone & \cfull & \cfull & 1 & \cfull\\
LiveStar~\citep{yang2025livestar} & M & V & \cpart & \cnone & \cnone & \cfull & \cfull & 1 & \cfull\\
MMDuet2~\citep{wang2025mmduet2} & M & V & \cpart & \cnone & \cnone & \cfull & \cfull & 1 & \cfull\\
ROMA~\citep{tian2026roma} & M & V & \cfull & \cnone & \cnone & \cfull & \cfull & 1 & \cfull\\
StreamOV~\citep{xie2026streamov} & M & V & \cfull & \cnone & \cnone & \cfull & \cfull & 1 & \cfull\\
\midrule
LiViBench~\citep{wang2026livibench} & B & L & \cfull & \cpart & \cpart & \cnone & \cnone & 1 & \cfull\\
Click-to-Ask~\citep{yu2026clicktoask} & S & L & \cpart & \cpart & \cpart & \cpart & \cpart & 1 & \cfull\\
VideoModerator~\citep{tang2021videomoderator} & S & L & \cfull & \cnone & \cpart & \cpart & \cnone & 1 & \cpart\\
\midrule
\rowcolor{oursbg}
\liveassistant\ourstar & B+M & L & \cfull & \cfull & \cfull & \cfull & \cfull & \textbf{2} & \cfull\\
\bottomrule
\end{tabular}%
}
\end{table}

\section{Method}
\label{sec:method}

\liveassistant learns a streaming policy over synchronized livestream trajectories for the decision problem in Section~\ref{sec:formulation}. A data engine converts replay media and platform logs into chunk aligned supervision. The policy is then optimized in two stages. MA-MSFT establishes the structured action protocol from reference histories. SM-GSPO continues optimization with self generated trajectories so that each action conditions later decisions on the policy history.

\begin{figure}[t]
    \centering
    \includegraphics[width=\linewidth]{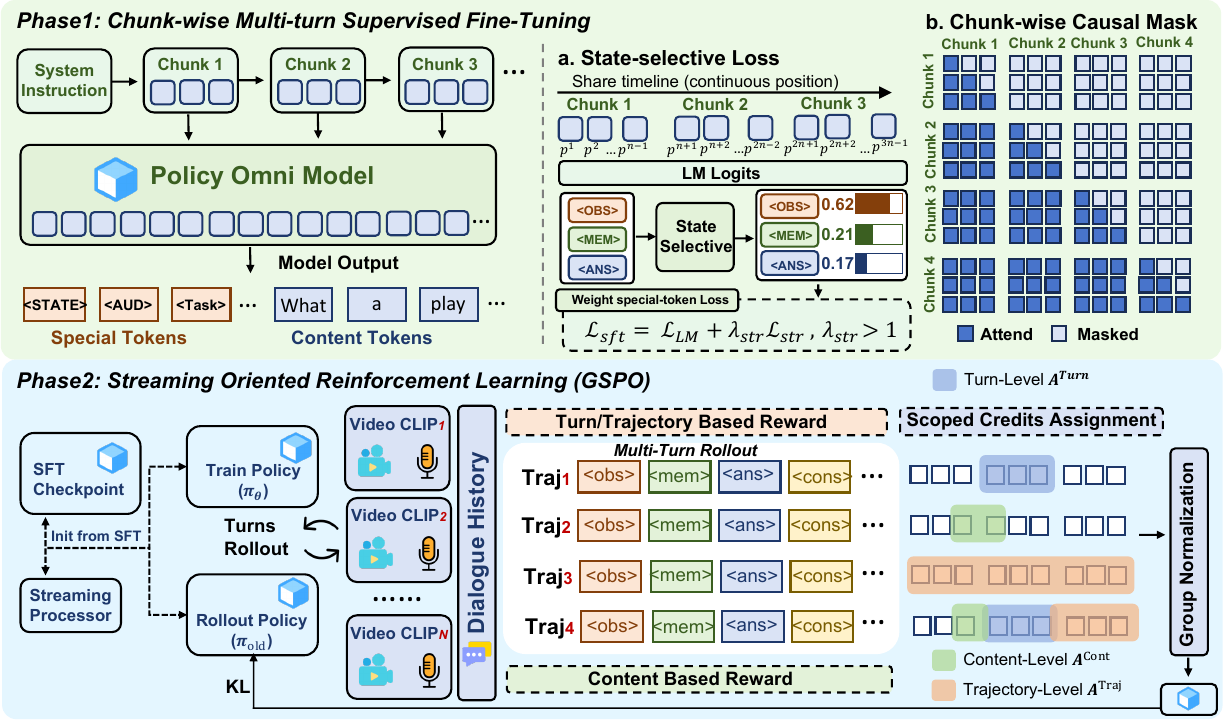}
    \caption{Overview of the \liveassistant data and training pipeline. A data engine constructs synchronized trajectories, MA-MSFT establishes the structured policy, and SM-GSPO continues optimization on self generated streaming trajectories.}
    \label{fig:method}
\end{figure}

\subsection{Livestream Trajectory Construction}
\label{sec:data_engine}

We convert real world recorded livestreams and synchronized platform logs into multiturn decision trajectories. The pipeline reconstructs continuous media, divides each stream into ten second decision intervals, and aligns native audio and video with comments, gifts, viewer dynamics, and room metadata. For each interval, Gemini 3.1 Pro receives the available stream context and produces a structured \textsc{OBS}, \textsc{MEM}, or \textsc{ANS} action following Equation~\ref{eq:action_protocol}. Automatic validation and human review enforce temporal consistency, valid recipient task combinations, and grounded outputs. Data are split by room before contiguous training windows are constructed.

The policy is instantiated with the thinker component of Qwen3-Omni~\citep{qwen2025qwen3omni}. Native audio and video are processed by the corresponding modality encoders. Synchronized platform signals are serialized as timestamped text. Media tokens, social signals, and previous assistant actions are arranged in temporal order so that one autoregressive policy can determine state, recipient, task, and response content.

\subsection{Marker Aware Multiturn Supervised Fine Tuning}
\label{sec:ma_msft}

The structured targets in Equation~\ref{eq:action_protocol} are asymmetric in length. An \textsc{OBS} decision may contribute one supervised token, while \textsc{MEM} and \textsc{ANS} contain many content tokens. Token averaged cross entropy therefore provides substantially more gradient to response wording than to the short structural markers that control system behavior. Let $y_{t,j}$ be the $j$th assistant token of action $y_t$, let $\chi_{t,j}\in\{0,1\}$ indicate whether the token is supervised, and let $\eta(y_{t,j})$ denote a marker aware weight. We retain the full vocabulary likelihood and increase the contribution of protocol markers:
\begin{equation}
    \mathcal{L}_{\mathrm{vocab}}
    =
    -
    \frac{
        \sum_{t,j}
        \chi_{t,j}\eta(y_{t,j})
        \log\pi_\theta(y_{t,j}\mid h_t,y_{t,<j})
    }{
        \sum_{t,j}
        \chi_{t,j}\eta(y_{t,j})
    }.
\label{eq:weighted_vocab}
\end{equation}
The weight $\eta(\cdot)$ distinguishes state, recipient, task, and ordinary content tokens, preserving open vocabulary generation while strengthening rare structural decisions.

Marker weighting still scores the state marker against the full vocabulary. We therefore form a restricted state distribution from the same logits,
\begin{equation}
    \pi^{\mathcal{S}}_{\theta}(z\mid h_t)
    =
    \operatorname{softmax}_{z\in\mathcal{S}}\xi_t(z),
\end{equation}
and apply a class weighted focal objective
\begin{equation}
    \mathcal{L}_{\mathrm{state}}
    =
    -\frac{
        \sum_{t=1}^{T}
        \beta_{z_t}
        \operatorname{sg}\!\left[(1-\pi^{\mathcal{S}}_{\theta}(z_t\mid h_t))^{\gamma}\right]
        \log \pi^{\mathcal{S}}_{\theta}(z_t\mid h_t)
    }{
        \sum_{t=1}^{T}\beta_{z_t}
    }.
\label{eq:state_candidate}
\end{equation}
Here $\beta_{z_t}$ compensates for state imbalance, $\gamma$ emphasizes uncertain state decisions~\citep{lin2017focal}, and $\operatorname{sg}$ stops the focal weight from becoming an optimization path. Restricted label token objectives have been used for language model classification~\citep{schick2021pet}. The complete supervised objective is
\begin{equation}
    \mathcal{L}_{\mathrm{MA\text{-}MSFT}}
    =
    \mathcal{L}_{\mathrm{vocab}}
    +
    \alpha\mathcal{L}_{\mathrm{state}}.
\end{equation}
MA-MSFT uses reference histories and establishes a strong structured policy before streaming reinforcement learning.

\subsection{Streaming Multiturn GSPO}
\label{sec:sm_gspo}

Deployment conditions every decision on the model actions from earlier turns. This creates an exposure mismatch with supervised training and allows early state or memory errors to affect the rest of a stream~\citep{bengio2015scheduledsampling}. SM-GSPO addresses this mismatch with self generated trajectories and complementary turn level and trajectory level credit.

\paragraph{On policy multiturn rollout.}
At each training step, the resident rollout model is synchronized with the current training model and generates a group of $G$ trajectories. Each turn conditions on the incoming stream and the model actions generated in previous turns:
\begin{equation}
    \widehat y_{g,t}
    \sim
    \mu\bigl(\cdot\mid x_{\leq t},\widehat y_{g,<t}\bigr),
\end{equation}
where $g\in\{1,\ldots,G\}$ indexes rollouts and $\mu$ is the behavior policy induced by decoding. The sampled trajectories remain fixed during optimization. Before the first optimizer update, the training model replays them to compute $\pi_{\theta_{\mathrm{old}}}$, which remains fixed across subsequent updates on the rollout batch.

\paragraph{Turn and trajectory rewards.}
Each turn receives a structural reward $\mathcal{R}^{\mathrm{str}}_{g,t}$ for format, state, recipient, task, and message count correctness. For a correctly predicted \textsc{ANS} state, a Qwen3-VL judge provides a content score $c_{g,t}\in[0,1]$. With a state shaping term $b_{g,t}$ and a format gate $f_{g,t}$, the turn and trajectory rewards are
\begin{equation}
    s_{g,t}
    =
    f_{g,t}
    \left(
        \mathcal{R}^{\mathrm{str}}_{g,t}
        +
        \lambda_c c_{g,t}
        +
        b_{g,t}
    \right),
    \qquad
    R_g
    =
    \sum_{t=1}^{T_g}\mathcal{R}^{\mathrm{str}}_{g,t}.
\label{eq:multiscale_rewards}
\end{equation}
The turn reward provides local feedback for each decision. The trajectory reward measures structural consistency across the complete rollout.

\paragraph{Scoped credit assignment.}
We standardize $\{s_{g,t}\}_{g=1}^{G}$ and $\{R_g\}_{g=1}^{G}$ within each rollout group to obtain the turn advantage $A_{g,t}$ and trajectory advantage $A_g$. Let $\mathcal{I}_{g,t}$ denote the assistant token positions generated at turn $t$. The turn advantage is assigned only to these tokens, with the GSPO ratio
\begin{equation}
    \rho_{g,t}
    =
    \exp\left[
        \frac{1}{|\mathcal{I}_{g,t}|}
        \sum_{j\in\mathcal{I}_{g,t}}
        \log
        \frac{
            \pi_\theta(\widehat y_{g,t,j}\mid H_{g,t,j})
        }{
            \pi_{\theta_{\mathrm{old}}}(\widehat y_{g,t,j}\mid H_{g,t,j})
        }
    \right],
\label{eq:turn_gspo_ratio}
\end{equation}
where $H_{g,t,j}$ is the complete causal prefix preceding token $j$. The trajectory advantage is assigned across all assistant tokens in rollout $g$ with the corresponding trajectory ratio $\rho_g$. This scope preserves local credit for each turn while retaining a signal for complete trajectory behavior.

\paragraph{Training objective.}
Using the clipped GSPO surrogate $\ell$~\citep{shao2025gspo}, the final objective combines turn level and trajectory level credit:
\begin{equation}
    \mathcal{L}_{\mathrm{SM\text{-}GSPO}}
    =
    \frac{1}{\sum_g T_g}
    \sum_{g,t}
    \ell(\rho_{g,t},A_{g,t})
    +
    \frac{1}{G}
    \sum_g
    \ell(\rho_g,A_g).
\label{eq:sm_gspo}
\end{equation}
The first term trains each turn from its local reward. The second term promotes consistent structured behavior across the rollout.

\subsection{Long Stream Inference}

At inference time, each chunk appends new media and platform tokens to a continuous streaming cache. The runtime retains the system instruction and the recent twelve turns at full resolution. Older audio and visual states are evicted, while compact \textsc{MEM} and text states preserve semantic continuity. This design follows the separation between dense recent perception and compressed long term memory used in streaming video systems~\citep{xu2026streamingvlm,liu2026thinkstream}.

\section{Experiments}
\label{sec:experiments}

The release evaluation focuses on the central claims of the paper. We study whether the policy makes the correct state decision, whether it routes actions to the appropriate participant, and how the two training stages contribute to the final policy. All benchmark inputs are processed chronologically with future observations hidden.

\subsection{Experimental Setup}

\paragraph{Training data and benchmark.}
Streams are split by room before model inputs are constructed. The optimization data use 115 rooms for MA-MSFT and eight disjoint rooms for SM-GSPO. Fourteen additional rooms are reserved for evaluation. Optimization samples are contiguous windows of five to twelve turns. Benchmark samples retain complete trajectories. The benchmark contains 275 clips and 13,812 unique ten second chunks. It spans nine content categories and seven languages and includes all three decision states, both recipient roles, and synchronized social signals.

\begin{table}[t]
\centering
\caption{Summary statistics of the release benchmark and optimization corpus. Counts follow room level partitioning and deduplicate repeated chunks introduced by overlapping multiturn windows.}
\label{tab:data_stats}
\small
\setlength{\tabcolsep}{6pt}
\renewcommand{\arraystretch}{1.1}
\begin{tabular}{@{}lrr@{}}
\toprule
\textbf{Statistic} & \textbf{Optimization} & \textbf{Benchmark}\\
\midrule
Livestream rooms & 123 & 14\\
Continuous clips & 2,049 & 275\\
Unique ten second chunks & 115,938 & 13,812\\
Duration in hours & 320.8 & 38.4\\
Content categories & 16 & 9\\
Language groups & 7 & 7\\
Comments & 4,124,223 & 625,119\\
Gifts & 104,146 & 19,217\\
\bottomrule
\end{tabular}
\end{table}

\paragraph{Baselines.}
We include two groups of baselines. Native streaming video models are evaluated with their released runtime behavior and provide a state decision score when their output protocol can be mapped to the benchmark. General multimodal models receive the unified livestream input and are prompted to emit the same structured action protocol as \liveassistant. The general baseline set includes Qwen3-VL, Qwen3.5, Qwen3-Omni, and InternVL3.5 variants~\citep{qwen2025qwen3vl,qwen2026qwen35,qwen2025qwen3omni,wang2025internvl35}.

\paragraph{Metrics.}
State All is the overall accuracy over \textsc{OBS}, \textsc{MEM}, and \textsc{ANS}. OBS, MEM, and ANS report per state recall. Recipient All is overall recipient accuracy on reference \textsc{ANS} turns, with Viewer and Host reporting role conditioned accuracy. Task is end to end task accuracy, with zero credit when the state decision is incorrect. Sem. is the semantic content score for the trained \liveassistant variants. The main comparison emphasizes the structured metrics that are shared across methods.

\paragraph{Implementation.}
\liveassistant is initialized from the thinker component of Qwen3-Omni 30B A3B~\citep{qwen2025qwen3omni}. MA-MSFT uses full parameter optimization for three epochs with a global batch size of 512, a peak learning rate of $1\times10^{-5}$, and a 32,768 token context. SM-GSPO starts from the MA-MSFT checkpoint and uses group size five, twelve turn rollouts, two optimizer updates per rollout batch, learning rate $1\times10^{-7}$, and temperature 0.5. Benchmark inference uses greedy decoding and a maximum generation length of 64 tokens per turn.

\subsection{Main Comparison}

Table~\ref{tab:main_results} reports representative streaming and general multimodal baselines under the same benchmark protocol. Native streaming models provide partial coverage of the structured protocol, while general multimodal models expose characteristic state and recipient biases under the unified evaluation. \liveassistant reaches 71.14 State All, 72.67 Recipient All, and 58.41 Task accuracy. The per state and per role columns show that the policy maintains substantial accuracy on \textsc{MEM} and host directed actions, which are the two most difficult parts of the benchmark.

\begin{table}[t]
\centering
\caption{Main comparison on the \liveassistant benchmark. NA denotes fields outside the native output protocol of a baseline or fields without evaluation for that model.}
\label{tab:main_results}
\scriptsize
\setlength{\tabcolsep}{2.2pt}
\renewcommand{\arraystretch}{1.08}
\resizebox{\linewidth}{!}{%
\begin{tabular}{@{}ll cccc ccc c c@{}}
\toprule
& & \multicolumn{4}{c}{\textbf{State decision}} & \multicolumn{3}{c}{\textbf{Recipient routing}} & \textbf{Action} & \textbf{Content}\\
\cmidrule(lr){3-6}\cmidrule(lr){7-9}\cmidrule(lr){10-10}\cmidrule(lr){11-11}
\textbf{Method} & \textbf{Input} & \textbf{All} & \textbf{OBS} & \textbf{MEM} & \textbf{ANS} & \textbf{All} & \textbf{Viewer} & \textbf{Host} & \textbf{Task} & \textbf{Sem.}\\
\midrule
\multicolumn{11}{@{}l}{\textit{Native streaming video models}}\\
Dispider~\citep{qian2025dispider} & V & 46.51 & NA & NA & NA & NA & NA & NA & NA & NA\\
LiveCC~\citep{chen2025livecc} & V & 27.06 & NA & NA & NA & NA & NA & NA & NA & NA\\
StreamingVLM~\citep{xu2026streamingvlm} & V & 2.17 & NA & NA & NA & NA & NA & NA & NA & NA\\
LiveStarPro~\citep{yang2026livestarpro} & V & 42.85 & NA & NA & NA & NA & NA & NA & NA & NA\\
MMDuet2~\citep{wang2025mmduet2} & V & 0.82 & NA & NA & NA & NA & NA & NA & NA & NA\\
\midrule
\multicolumn{11}{@{}l}{\textit{General multimodal models under unified streaming evaluation}}\\
Qwen3-VL-4B~\citep{qwen2025qwen3vl} & V+C+G & 42.17 & 32.42 & 23.48 & 79.06 & 66.87 & 75.86 & 11.45 & 41.94 & NA\\
Qwen3-VL-8B~\citep{qwen2025qwen3vl} & V+C+G & 52.51 & 49.09 & 33.39 & 78.05 & 50.34 & 52.61 & 36.35 & 38.24 & NA\\
Qwen3.5-9B~\citep{qwen2026qwen35} & V+C+G & 43.22 & 25.71 & 28.69 & 90.44 & 68.33 & 74.43 & 30.72 & 46.43 & NA\\
Qwen3.5-27B~\citep{qwen2026qwen35} & V+C+G & 43.30 & 23.79 & 35.07 & 87.92 & 71.05 & 76.48 & 37.55 & 40.12 & NA\\
Qwen3-Omni-30B-A3B~\citep{qwen2025qwen3omni} & A+V+C+G & 32.59 & 14.11 & 3.73 & 96.12 & 48.65 & 47.76 & 54.05 & 49.23 & NA\\
InternVL3.5-8B~\citep{wang2025internvl35} & V+C+G & 58.05 & 78.50 & 27.00 & 51.02 & 29.21 & 29.30 & 28.66 & 27.34 & NA\\
InternVL3.5-14B~\citep{wang2025internvl35} & V+C+G & 33.13 & 16.92 & 3.44 & 93.09 & 49.36 & 49.33 & 49.50 & 30.51 & NA\\
\midrule
\rowcolor{oursbg}
\liveassistant\ourstar & A+V+C+G & \textbf{71.14} & 78.84 & \textbf{48.20} & 79.26 & \textbf{72.67} & 75.57 & \textbf{56.71} & \textbf{58.41} & \textbf{81.78}\\
\bottomrule
\end{tabular}%
}
\vspace{2pt}

\raggedright\footnotesize A denotes native audio, V video, C comments, and G gifts. Streaming baselines expose only the fields supported by their released output protocol.
\end{table}

\subsection{Contribution of the Training Stages}

Table~\ref{tab:training_stages} isolates the optimization stages under the same backbone and input protocol. MA-MSFT improves structured state learning and provides the checkpoint used by reinforcement learning. Turn level GSPO increases recipient and task accuracy. Full SM-GSPO gives the strongest joint structured result and raises MEM recall to 48.20, Recipient All to 72.67, and Task to 58.41.

\begin{table}[t]
\centering
\caption{Training stage comparison. The table keeps the structured metrics that show the contribution of the optimization stages.}
\label{tab:training_stages}
\small
\setlength{\tabcolsep}{5.3pt}
\renewcommand{\arraystretch}{1.1}
\begin{tabular}{@{}lrrrr@{}}
\toprule
\textbf{Training strategy} & \textbf{State All} & \textbf{MEM} & \textbf{Recipient All} & \textbf{Task}\\
\midrule
Ordinary multiturn SFT & 69.36 & 35.99 & 63.54 & 51.12\\
MA-MSFT & 70.05 & 36.50 & 66.42 & 52.55\\
MA-MSFT + turn level GSPO & 69.98 & 28.52 & 68.81 & 54.46\\
\rowcolor{oursbg}
MA-MSFT + SM-GSPO & \textbf{71.14} & \textbf{48.20} & \textbf{72.67} & \textbf{58.41}\\
\bottomrule
\end{tabular}
\end{table}

\section{Related Work}

\paragraph{Proactive and online streaming video understanding.}
Online video models process content as it unfolds. Early systems established causal frame processing and persistent visual memory~\citep{zhang2024flashvstream,chen2024videollmonline}. Benchmarks then evaluated temporal grounding, current understanding, and responses that depend on accumulated evidence~\citep{huang2025ovbench,li2025ovobench,yang2025svbench,lin2024streamingbench,wang2025omnimmi}. Recent systems learn response timing and selective activation. Dispider separates perception, decision, and reaction. StreamBridge learns activation. LiveStar models response and silence decisions. StreamReady studies evidence sufficiency at emission time~\citep{qian2025dispider,wang2025streambridge,yang2025livestar,azad2026streamready}. MMDuet2 and StreamPro use multiturn reinforcement learning for timing and restraint~\citep{wang2025mmduet2,li2026streampro}. ROMA, StreamOV, LiveVLM, StreamingVLM, and LiveStarPro further develop native audio visual processing, memory, and bounded streaming context~\citep{tian2026roma,xie2026streamov,ning2025livevlm,xu2026streamingvlm,yang2026livestarpro}. This line provides the streaming perception and timing machinery used by livestream assistance.

\paragraph{Livestream as a social and platform environment.}
Livestream research models comments and platform activity as first class signals. LiveBot and LiveChat generate comments from multimodal context~\citep{ma2019livebot,lalanne2023livechat}. LiViBench combines native audio, speech, and real time comments for livestream question answering~\citep{wang2026livibench}. KuaiLive and KuaiLive M3 expose temporally evolving content, interaction, and behavioral feedback for recommendation research~\citep{qu2026kuailive,guo2026kuailivem3}. VideoModerator and recent industrial systems use multimodal evidence for livestream safety workflows~\citep{tang2021videomoderator,yew2026dynamicmoderation,ye2026failuretaxonomy,qiao2026dejavu}. Click-to-Ask introduces streaming memory for viewer inquiries~\citep{yu2026clicktoask}. These systems demonstrate the value of social and platform signals and motivate a unified assistant that selects among multiple possible actions.

\paragraph{Mixed initiative and multi party interaction.}
Mixed initiative systems study how control is shared between users and automated agents~\citep{horvitz1999principles,allen1999mixedinitiative}. Work on interruption models the timing of assistance as a decision over user attention~\citep{horvitz2003interruption,fogarty2005interruptibility}. Multi party dialogue adds an explicit recipient dimension through turn taking, addressee recognition, and group participation~\citep{sacks1974turntaking,jovanovic2006addressee,bohus2009models,houde2025groupagent,inoue2025addressee}. Recent proactive dialogue research extends initiative taking to language agents~\citep{deng2023prompting,deng2025proactive}. Livestream assistance instantiates this decision vocabulary over a native multimodal social stream with asynchronously emerging needs.

\paragraph{Multiturn reinforcement learning.}
Group relative optimization provides critic free policy learning from sampled groups, while GSPO uses sequence level ratios for language model policies~\citep{shao2024deepseekmath,shao2025gspo}. Turn level reward design gives dense credit inside long interactions~\citep{wei2025turnlevel}. SM-GSPO combines turn level credit for local activation and routing with trajectory level credit for complete streaming behavior. This formulation matches the causal structure of livestream interaction, where each generated action becomes part of the context for subsequent decisions.

\section{Conclusion}

We formulate livestream assistance as mixed initiative, role conditioned decision making over a multi party social stream. The model receives native audio, video, comments, gifts, and room dynamics, then resolves whether to act, when to act, whom to address, and what to communicate through one structured autoregressive policy. A trajectory data engine provides causal supervision from real livestreams. MA-MSFT strengthens sparse protocol decisions, and SM-GSPO trains on self generated trajectories with scoped turn and trajectory credit. The release benchmark shows strong gains in state decision, recipient routing, and task selection across representative streaming and general multimodal baselines. The resulting framework provides a concrete foundation for assistants that participate selectively in live social environments.

\paragraph{Limitations and broader impact.}
The current benchmark covers nine content categories and seven languages, and performance on unseen domains and platforms requires further study. Host directed actions are less frequent than viewer directed actions, which makes host routing the more data limited case. Content quality uses an automatic judge and inherits the limitations of model based evaluation. The ten second decision interval also limits the temporal precision of interventions. Production deployment would additionally require consent, privacy protection, role specific controls, and auditing of routing and activation errors.

\bibliographystyle{unsrtnat}
\bibliography{refs}

@article{chen2024videollmonline,
  title={VideoLLM-online: Online Video Large Language Model for Streaming Video},
  author={Joya Chen and Zhaoyang Lv and Shiwei Wu and Kevin Qinghong Lin and Chenan Song and Difei Gao and Jia-Wei Liu and Ziteng Gao and Dongxing Mao and Mike Zheng Shou},
  journal={2024 IEEE/CVF Conference on Computer Vision and Pattern Recognition (CVPR)},
  year={2024},
  pages={18407-18418},
  url={https://api.semanticscholar.org/CorpusID:270560262}
}

@article{zhang2024flashvstream,
  title         = {{Flash-VStream}: Memory-Based Real-Time Understanding for Long Video Streams},
  author        = {Zhang, Haoji and Wang, Yiqin and Tang, Yansong and Liu, Yong and Feng, Jiashi and Dai, Jifeng and Jin, Xiaojie},
  journal       = {arXiv preprint arXiv:2406.08085
        
        
        
        
        
        },
  year          = {2024},
  eprint        = {2406.08085},
  archivePrefix = {arXiv}
}

@article{lin2024streamingbench,
  title         = {{StreamingBench}: Assessing the Gap for MLLMs to Achieve Streaming Video Understanding},
  author        = {Lin, Junming and Fang, Zheng and Chen, Chi and Wan, Zihao and Luo, Fuwen and Li, Peng and Liu, Yang and Sun, Maosong},
  journal       = {arXiv preprint arXiv:2411.03628
        
        
        
        
        
        },
  year          = {2024},
  eprint        = {2411.03628},
  archivePrefix = {arXiv}
}

@article{huang2025ovbench,
  title         = {Online Video Understanding: {OVBench} and {VideoChat-Online}},
  author        = {Huang, Zhenpeng and Li, Xinhao and Li, Jiaqi and Wang, Jing and Zeng, Xiangyu and Liang, Cheng and Wu, Tao and Chen, Xi and Li, Liang and Wang, Limin},
  journal       = {Proceedings of the IEEE/CVF Conference on Computer Vision and Pattern Recognition},
  year          = {2025},
  eprint        = {2501.00584},
  archivePrefix = {arXiv}
}

@article{li2025ovobench,
  title         = {{OVO-Bench}: How Far is Your Video-LLMs from Real-World Online Video Understanding?},
  author        = {Li, Yifei and Niu, Junbo and Miao, Ziyang and Ge, Chunjiang and Zhou, Yuanhang and He, Qihao and Dong, Xiaoyi and Duan, Haodong and Ding, Shuangrui and Qian, Rui and Zhang, Pan and Zang, Yuhang and Cao, Yuhang and He, Conghui and Wang, Jiaqi},
  journal       = {Proceedings of the IEEE/CVF Conference on Computer Vision and Pattern Recognition},
  year          = {2025},
  eprint        = {2501.05510},
  archivePrefix = {arXiv}
}

@article{yang2025svbench,
  title         = {{SVBench}: A Benchmark with Temporal Multi-Turn Dialogues for Streaming Video Understanding},
  author        = {Yang, Zhenyu and Hu, Yuhang and Du, Zemin and Xue, Dizhan and Qian, Shengsheng and Wu, Jiahong and Yang, Fan and Dong, Weiming and Xu, Changsheng},
  journal       = {International Conference on Learning Representations},
  year          = {2025},
  eprint        = {2502.10810},
  archivePrefix = {arXiv}
}

@article{wang2025omnimmi,
  title         = {{OmniMMI}: A Comprehensive Multi-modal Interaction Benchmark in Streaming Video Contexts},
  author        = {Wang, Yuxuan and Wang, Yueqian and Chen, Bo and Wu, Tong and Zhao, Dongyan and Zheng, Zilong},
  journal       = {Proceedings of the IEEE/CVF Conference on Computer Vision and Pattern Recognition},
  year          = {2025},
  eprint        = {2503.22952},
  archivePrefix = {arXiv}
}

@article{qian2025dispider,
  title         = {Dispider: Enabling Video LLMs with Active Real-Time Interaction via Disentangled Perception, Decision, and Reaction},
  author        = {Qian, Rui and Ding, Shuangrui and Dong, Xiaoyi and Zhang, Pan and Zang, Yuhang and Cao, Yuhang and Lin, Dahua and Wang, Jiaqi},
  journal       = {arXiv preprint arXiv:2501.03218
        
        },
  year          = {2025},
  eprint        = {2501.03218},
  archivePrefix = {arXiv}
}

@article{chen2025livecc,
  title         = {{LiveCC}: Learning Video LLM with Streaming Speech Transcription at Scale},
  author        = {Chen, Joya and Zeng, Ziyun and Lin, Yiqi and Li, Wei and Ma, Zejun and Shou, Mike Zheng},
  journal       = {Proceedings of the IEEE/CVF Conference on Computer Vision and Pattern Recognition},
  year          = {2025},
  eprint        = {2504.16030},
  archivePrefix = {arXiv}
}

@article{wang2025streambridge,
  title         = {StreamBridge: Turning Your Offline Video Large Language Model into a Proactive Streaming Assistant},
  author        = {Wang, Haibo and Feng, Bo and Lai, Zhengfeng and Xu, Mingze and Li, Shiyu and Ge, Weifeng and Dehghan, Afshin and Cao, Meng and Huang, Ping},
  journal       = {Advances in Neural Information Processing Systems},
  year          = {2025},
  eprint        = {2505.05467},
  archivePrefix = {arXiv}
}

@article{ning2025livevlm,
  title         = {{LiveVLM}: Efficient Online Video Understanding via Streaming-Oriented KV Cache and Retrieval},
  author        = {Ning, Zhenyu and Liu, Guangda and Jin, Qihao and Li, Chengwei and Ding, Wenchao and Guo, Minyi and Zhao, Jieru},
  journal       = {arXiv preprint arXiv:2505.15269
        
        
        
        },
  year          = {2025},
  eprint        = {2505.15269},
  archivePrefix = {arXiv}
}

@article{yang2025livestar,
  title         = {LiveStar: Live Streaming Assistant for Real-World Online Video Understanding},
  author        = {Yang, Zhenyu and Zhang, Kairui and Hu, Yuhang and Wang, Bing and Qian, Shengsheng and Wen, Bin and Yang, Fan and Gao, Tingting and Dong, Weiming and Xu, Changsheng},
  journal       = {Advances in Neural Information Processing Systems},
  year          = {2025},
  eprint        = {2511.05299},
  archivePrefix = {arXiv}
}

@article{wang2025mmduet2,
  title         = {{MMDuet2}: Enhancing Proactive Interaction of Video MLLMs with Multi-Turn Reinforcement Learning},
  author        = {Wang, Yueqian and Liu, Songxiang and Wang, Disong and Xu, Nuo and Wan, Guanglu and Zhang, Huishuai and Zhao, Dongyan},
  journal       = {arXiv preprint arXiv:2512.06810
        
        },
  year          = {2025},
  eprint        = {2512.06810},
  archivePrefix = {arXiv}
}

@article{xu2026streamingvlm,
  title         = {{StreamingVLM}: Real-Time Understanding for Infinite Video Streams},
  author = {Xu, Ruyi and Xiao, Guangxuan and Chen, Yukang and He, Liuning and Peng, Kelly and Lu, Yao and Han, Song},
  journal       = {International Conference on Learning Representations},
  year          = {2026},
  eprint        = {2510.09608},
  archivePrefix = {arXiv}
}

@article{azad2026streamready,
  title         = {StreamReady: Learning What to Answer and When in Long Streaming Videos},
  author        = {Azad, Shehreen and Vineet, Vibhav and Rawat, Yogesh Singh},
  journal       = {Proceedings of the IEEE/CVF Conference on Computer Vision and Pattern Recognition},
  year          = {2026},
  eprint        = {2603.08620},
  archivePrefix = {arXiv}
}

@article{tian2026roma,
  title         = {{ROMA}: Real-time Omni-Multimodal Assistant with Interactive Streaming Understanding},
  author        = {Tian, Xueyun and Li, Wei and Xu, Bingbing and Dong, Heng and Wang, Yuanzhuo and Shen, Huawei},
  journal       = {arXiv preprint arXiv:2601.10323
        
        },
  year          = {2026},
  eprint        = {2601.10323},
  archivePrefix = {arXiv}
}

@article{li2026streampro,
  title         = {StreamPro: From Reactive Perception to Proactive Decision-Making in Streaming Video},
  author        = {Li, Ao and Xiao, Zihan and Yue, Zihao and Xu, Boshen and Yao, Linli and Li, Jiaze and Fu, Pei and Ju, Jianzhong and Luan, Jian and Jin, Qin},
  journal       = {arXiv preprint arXiv:2605.16381},
  year          = {2026},
  eprint        = {2605.16381},
  archivePrefix = {arXiv}
}

@article{xie2026streamov,
  title         = {StreamOV: Streaming Omni-Video Understanding via Evidence-Guided Memory and Response Triggering},
  author        = {Xie, Ming and Huang, Zizheng and Tan, Xudong and Wang, Chao and Zeng, Xiangyu and Wu, Wenxiao and Chen, Tao and Wang, Limin and Fu, Yanwei},
  journal       = {arXiv preprint arXiv:2605.25621},
  year          = {2026},
  eprint        = {2605.25621},
  archivePrefix = {arXiv}
}

@article{yang2026livestarpro,
  title         = {LiveStarPro: Proactive Streaming Video Understanding with Hierarchical Memory for Long-Horizon Streams},
  author        = {Yang, Zhenyu and Zhang, Kairui and Wang, Bing and Qian, Shengsheng and Xu, Changsheng},
  journal       = {arXiv preprint arXiv:2606.17798},
  year          = {2026},
  eprint        = {2606.17798},
  archivePrefix = {arXiv}
}

@article{ma2019livebot,
  title   = {LiveBot: Generating Live Video Comments Based on Visual and Textual Contexts},
  author  = {Ma, Shuming and Cui, Lei and Dai, Damai and Wei, Furu and Sun, Xu},
  journal = {Proceedings of the AAAI Conference on Artificial Intelligence},
  volume  = {33},
  number  = {01},
  pages   = {6810--6817},
  year    = {2019},
  doi     = {10.1609/aaai.v33i01.33016810}
}

@article{lalanne2023livechat,
  title         = {LiveChat: Video Comment Generation from Audio-Visual Multimodal Contexts},
  author        = {Lalanne, Julien and Bournet, Raphael and Yu, Yi},
  journal       = {arXiv preprint arXiv:2311.12826
        
        },
  year          = {2023},
  eprint        = {2311.12826},
  archivePrefix = {arXiv}
}

@article{wang2026livibench,
  title   = {LiViBench: An Omnimodal Benchmark for Interactive Livestream Video Understanding},
  author  = {Wang, Xiaodong and Huang, Langling and Wu, Zhirong and Zhao, Xu and Xu, Teng and Xia, Xuhong and Peng, Peixi},
  journal = {Proceedings of the AAAI Conference on Artificial Intelligence},
  volume  = {40},
  number  = {31},
  pages   = {26517--26525},
  year    = {2026},
  doi     = {10.1609/aaai.v40i31.39859}
}

@inproceedings{yu2026clicktoask,
  title     = {Click-to-Ask: An AI Live Streaming Assistant with Offline Copywriting and Online Interactive QA},
  author    = {Yu, Ruizhi and Zhong, Keyang and Liu, Peng and Wu, Qi and Zhang, Haoran and Zhang, Yanhao and Chen, Chen and Lu, Haonan},
  booktitle = {Companion Proceedings of the ACM Web Conference 2026},
  pages     = {192--195},
  year      = {2026},
  doi       = {10.1145/3774905.3793134}
}

@inproceedings{qu2026kuailive,
  title     = {KuaiLive: A Real-time Interactive Dataset for Live Streaming Recommendation},
  author    = {Qu, Changle and Dai, Sunhao and Guo, Ke and Zhang, Xiao and Zhao, Liqin and Wang, Shijun and Niu, Yanan and Hu, Lantao and Li, Han and Xu, Jun},
  booktitle = {Proceedings of the 49th International ACM SIGIR Conference on Research and Development in Information Retrieval},
  pages     = {3356--3364},
  year      = {2026},
  doi       = {10.1145/3805712.3808587}
}

@article{guo2026kuailivem3,
  title         = {KuaiLive-M3: A Multi-Modal, Multi-Domain, and Multi-Feedback Dataset for Live Streaming Recommendation},
  author        = {Guo, Ke and Qu, Changle and Cheng, Jiayaqi and Zhang, Xiao and Wang, Shijun and Zhang, Xiaoyu and Wang, Xueliang and Zhang, Le and Hu, Lantao and Xu, Jun},
  journal       = {arXiv preprint arXiv:2607.24862
        
        },
  year          = {2026},
  eprint        = {2607.24862},
  archivePrefix = {arXiv}
}

@article{tang2021videomoderator,
  title         = {VideoModerator: A Risk-Aware Framework for Multimodal Video Moderation in E-Commerce},
  author        = {Tang, Tan and Wu, Yanhong and Yu, Lingyun and Li, Yuhong and Wu, Yingcai},
  journal       = {arXiv preprint arXiv:2109.03479
        
        },
  year          = {2021},
  eprint        = {2109.03479},
  archivePrefix = {arXiv}
}

@article{yew2026dynamicmoderation,
  title         = {Dynamic Content Moderation in Livestreams: Combining Supervised Classification with MLLM-Boosted Similarity Matching},
  author        = {Yew, Wei Chee and Xu, Hailun and Saha, Sanjay and Fan, Xiaotian and Ong, Hiok Hian and Wang, David Yuchen and Sarkar, Kanchan and Yang, Zhenheng and Guan, Danhui},
  journal       = {arXiv preprint arXiv:2512.03553
        
        },
  year          = {2026},
  eprint        = {2512.03553},
  archivePrefix = {arXiv}
}

@article{ye2026failuretaxonomy,
  title         = {From Failure Taxonomy to Intervention: A Diagnostic Methodology for Industry-Scale AVLM in Video and Live-Streaming Platform Moderation},
  author        = {Ye, Shuchang and Yu, Jinqiang and Xiao, Zhujun and Kong, Yajing and Lin, Yist Y. and Ma, Yang and Liu, Jiaxi and Xu, Xiaolei and Yu, Zheng},
  journal       = {arXiv preprint arXiv:2606.30059
        
        },
  year          = {2026},
  eprint        = {2606.30059},
  archivePrefix = {arXiv}
}

@article{qiao2026dejavu,
  title         = {Deja Vu in Plots: Leveraging Cross-Session Evidence with Retrieval-Augmented LLMs for Live Streaming Risk Assessment},
  author        = {Qiao, Yiran and Ao, Xiang and Chen, Jing and Liu, Yang and Zhong, Qiwei and He, Qing},
  journal       = {arXiv preprint arXiv:2601.16027
        
        
        
        
        
        },
  year          = {2026},
  eprint        = {2601.16027},
  archivePrefix = {arXiv}
}

@article{qwen2025qwen3omni,
  title         = {{Qwen3-Omni} Technical Report},
  author        = {Xu, Jin and Guo, Zhifang and Hu, Hangrui and Chu, Yunfei and Wang, Xiong and He, Jinzheng and Wang, Yuxuan and Shi, Xian and He, Ting and Zhu, Xinfa and Lv, Yuanjun and Wang, Yongqi and Guo, Dake and Wang, He and Ma, Linhan and Zhang, Pei and Zhang, Xinyu and Hao, Hongkun and Guo, Zishan and Yang, Baosong and Zhang, Bin and Ma, Ziyang and Wei, Xipin and Bai, Shuai and Chen, Keqin and Liu, Xuejing and Wang, Peng and Yang, Mingkun and Liu, Dayiheng and Ren, Xingzhang and Zheng, Bo and Men, Rui and Zhou, Fan and Yu, Bowen and Yang, Jianxin and Yu, Le and Zhou, Jingren and Lin, Junyang},
  journal       = {arXiv preprint arXiv:2509.17765
        
        
        
        
        
        },
  year          = {2025},
  eprint        = {2509.17765},
  archivePrefix = {arXiv}
}

@article{shao2025gspo,
  title         = {Group Sequence Policy Optimization},
  author        = {Zheng, Chujie and Liu, Shixuan and Li, Mingze and Chen, Xiong-Hui and Yu, Bowen and Gao, Chang and Dang, Kai and Liu, Yuqiong and Men, Rui and Yang, An and Zhou, Jingren and Lin, Junyang},
  journal       = {arXiv preprint arXiv:2507.18071
        
        
        
        
        
        },
  year          = {2025},
  eprint        = {2507.18071},
  archivePrefix = {arXiv}
}

@inproceedings{lin2017focal,
  title     = {Focal Loss for Dense Object Detection},
  author    = {Lin, Tsung-Yi and Goyal, Priya and Girshick, Ross and He, Kaiming and Doll{\'a}r, Piotr},
  booktitle = {Proceedings of the IEEE International Conference on Computer Vision},
  year      = {2017}
}

@inproceedings{schick2021pet,
  title     = {Exploiting Cloze-Questions for Few-Shot Text Classification and Natural Language Inference},
  author    = {Schick, Timo and Sch{\"u}tze, Hinrich},
  booktitle = {Proceedings of the 16th Conference of the European Chapter of the Association for Computational Linguistics},
  year      = {2021},
  eprint    = {2001.07676},
  archivePrefix = {arXiv}
}

@inproceedings{bengio2015scheduledsampling,
  title     = {Scheduled Sampling for Sequence Prediction with Recurrent Neural Networks},
  author    = {Bengio, Samy and Vinyals, Oriol and Jaitly, Navdeep and Shazeer, Noam},
  booktitle = {Advances in Neural Information Processing Systems},
  year      = {2015}
}

@article{shao2024deepseekmath,
  title         = {{DeepSeekMath}: Pushing the Limits of Mathematical Reasoning in Open Language Models},
  author        = {Shao, Zhihong and Wang, Peiyi and Zhu, Qihao and Xu, Runxin and Song, Junxiao and Bi, Xiao and Zhang, Haowei and Zhang, Mingchuan and Li, Y. K. and Wu, Y. and Guo, Daya},
  journal       = {arXiv preprint arXiv:2402.03300
        
        
        
        
        
        },
  year          = {2024},
  eprint        = {2402.03300},
  archivePrefix = {arXiv}
}

@article{liu2026thinkstream,
  title         = {Thinking in Streaming Video},
  author        = {Liu, Zikang and Guo, Longteng and Li, Handong and Zhen, Ru and He, Xingjian and Ji, Ruyi and Ren, Xiaoming and Zhang, Yanhao and Lu, Haonan and Liu, Jing},
  journal       = {arXiv preprint arXiv:2603.12938
        
        
        
        
        
        
        
        },
  year          = {2026},
  eprint        = {2603.12938},
  archivePrefix = {arXiv}
}

@inproceedings{horvitz1999principles,
  title     = {Principles of Mixed-Initiative User Interfaces},
  author    = {Horvitz, Eric},
  booktitle = {Proceedings of the SIGCHI Conference on Human Factors in Computing Systems (CHI)},
  pages     = {159--166},
  year      = {1999}
}

@article{allen1999mixedinitiative,
  title   = {Mixed-Initiative Interaction},
  author  = {Allen, James E. and Guinn, Curry I. and Horvitz, Eric},
  journal = {IEEE Intelligent Systems and their Applications},
  volume  = {14},
  number  = {5},
  pages   = {14--23},
  year    = {1999}
}

@inproceedings{horvitz2003interruption,
  title     = {Learning and Reasoning about Interruption},
  author    = {Horvitz, Eric and Apacible, Johnson},
  booktitle = {Proceedings of the 5th International Conference on Multimodal Interfaces (ICMI)},
  pages     = {20--27},
  year      = {2003}
}

@article{fogarty2005interruptibility,
  title   = {Predicting Human Interruptibility with Sensors},
  author  = {Fogarty, James and Hudson, Scott E. and Atkeson, Christopher G. and Avrahami, Daniel and Forlizzi, Jodi and Kiesler, Sara and Lee, Johnny C. and Yang, Jie},
  journal = {ACM Transactions on Computer-Human Interaction (TOCHI)},
  volume  = {12},
  number  = {1},
  pages   = {119--146},
  year    = {2005}
}

@inproceedings{deng2023prompting,
  title     = {Prompting and Evaluating Large Language Models for Proactive Dialogues: Clarification, Target-guided, and Non-collaboration},
  author    = {Deng, Yang and Liao, Lizi and Chen, Liang and Wang, Hongru and Lei, Wenqiang and Chua, Tat-Seng},
  booktitle = {Findings of the Association for Computational Linguistics: EMNLP},
  year      = {2023}
}

@inproceedings{deng2025proactive,
  title     = {A Survey on Proactive Dialogue Systems: Problems, Methods, and Prospects},
  author    = {Deng, Yang and Lei, Wenqiang and Lam, Wai and Chua, Tat-Seng},
  booktitle = {Proceedings of the Thirty-Second International Joint Conference on Artificial Intelligence},
  pages     = {6583--6591},
  year      = {2023},
  doi       = {10.24963/ijcai.2023/738}
}

@inproceedings{houde2025groupagent,
  title     = {Controlling {AI} Agent Participation in Group Conversations: A Human-Centered Approach},
  author    = {Houde, Stephanie and Brimijoin, Kristina and Muller, Michael and Ross, Steven I. and Silva Moran, Dario Andres and Gonzalez, Gabriel Enrique and Kunde, Siya and Foreman, Morgan A. and Weisz, Justin D.},
  booktitle = {Proceedings of the 30th International Conference on Intelligent User Interfaces (IUI)},
  pages     = {390--408},
  year      = {2025},
  doi       = {10.1145/3708359.3712089},
        
        
}

@article{inoue2025addressee,
  title   = {An {LLM} Benchmark for Addressee Recognition in Multi-modal Multi-party Dialogue},
  author  = {Inoue, Koji and Lala, Divesh and Elmers, Mikey and Ochi, Keiko and Kawahara, Tatsuya},
  journal = {arXiv preprint arXiv:2501.16643
        
        
        
        
        
        
        
        
        
        
        
        },
  year    = {2025}
}

@article{wei2025turnlevel,
  title   = {Reinforcing Multi-Turn Reasoning in {LLM} Agents via Turn-Level Reward Design},
  author  = {Wei, Quan and Zeng, Siliang and Li, Chenliang and Brown, William and Frunza, Oana and Deng, Wei and Schneider, Anderson and Nevmyvaka, Yuriy and Zhao, Yang Katie and Garcia, Alfredo and Hong, Mingyi},
  journal = {arXiv preprint arXiv:2505.11821
        
        
        
        
        
        
        
        
        
        
        
        
        
        
        
        
        
        
        
        },
  year    = {2025}
}

@inproceedings{bohus2009models,
  title     = {Models for Multiparty Engagement in Open-World Dialog},
  author    = {Bohus, Dan and Horvitz, Eric},
  booktitle = {Proceedings of the SIGDIAL 2009 Conference},
  pages     = {225--234},
  year      = {2009},
  address   = {London, UK},
  publisher = {Association for Computational Linguistics},
  doi       = {10.3115/1708376.1708409}
}

@article{sacks1974turntaking,
  title   = {A Simplest Systematics for the Organization of Turn-Taking for Conversation},
  author  = {Sacks, Harvey and Schegloff, Emanuel A. and Jefferson, Gail},
  journal = {Language},
  volume  = {50},
  number  = {4},
  pages   = {696--735},
  year    = {1974},
  doi     = {10.1353/lan.1974.0010}
}

@inproceedings{jovanovic2006addressee,
  title     = {Addressee Identification in Face-to-Face Meetings},
  author    = {Jovanovi{\'c}, Nataša and op den Akker, Rieks and Nijholt, Anton},
  booktitle = {Proceedings of the 11th Conference of the European Chapter of the Association for Computational Linguistics (EACL)},
  year      = {2006},
  address   = {Trento, Italy},
  publisher = {Association for Computational Linguistics}
}

@article{qwen2025qwen3vl,
  title         = {{Qwen3-VL} Technical Report},
  author        = {{Qwen Team}},
  journal       = {arXiv preprint arXiv:2511.21631
        
        
        
        
        
        
        
        
        
        },
  year          = {2025},
  eprint        = {2511.21631},
  archivePrefix = {arXiv}
}

@article{wang2025internvl35,
  title         = {{InternVL3.5}: Advancing Open-Source Multimodal Models in Versatility, Reasoning, and Efficiency},
  author        = {Wang, Weiyun and Gao, Zhangwei and Gu, Lixin and Pu, Hengjun and Cui, Long and Wei, Xingguang and Liu, Zhaoyang and Jing, Linglin and Ye, Shenglong and Shao, Jie and others},
  journal       = {arXiv preprint arXiv:2508.18265
        
        
        
        
        
        
        
        
        
        
        
        },
  year          = {2025},
  eprint        = {2508.18265},
  archivePrefix = {arXiv}
}

@misc{qwen2026qwen35,
    title = {Qwen3.5: Towards Native Multimodal Agents},
    url = {https://qwen.ai/blog?id=qwen3.5},
    author = {Qwen Team},
    month = {February},
    year = {2026}
}

\end{document}